\documentclass[a4paper, 10pt, conference]{IEEEtran}
\IEEEoverridecommandlockouts

\usepackage{subcaption}
\usepackage{float}
\usepackage{graphicx}
\usepackage{comment}
\usepackage{cite}
\usepackage{amsmath,amssymb,amsfonts,bm}
\usepackage{algorithmic}
\usepackage{textcomp}
\usepackage{xcolor}
\def\BibTeX{{\rm B\kern-.05em{\sc i\kern-.025em b}\kern-.08em
    T\kern-.1667em\lower.7ex\hbox{E}\kern-.125emX}}
\usepackage{booktabs}

\title{Exploring the Task-agnostic Trait of Self-supervised Learning in the Context of Detecting Mental Disorders}

\author{
Rohan Kumar Gupta and
Rohit Sinha

\\Department of Electronics and Electrical Engineering\\
Indian Institute of Technology Guwahati, Guwahati, India-781039\\
E-mail: {\{rohan\_kumar, rsinha\}@iitg.ac.in}}

\begin{document}

\maketitle

\begin{abstract}
 Self-supervised learning (SSL) has been investigated to generate task-agnostic representations across various domains. However, such investigation has not been conducted for detecting multiple mental disorders. The rationale behind the existence of a task-agnostic representation lies in the overlapping symptoms among multiple mental disorders. Consequently, the behavioural data collected for mental health assessment may carry a mixed bag of attributes related to multiple disorders. Motivated by that, in this study, we explore a task-agnostic representation derived through SSL in the context of detecting major depressive disorder (MDD) and post-traumatic stress disorder (PTSD) using audio and video data collected during interactive sessions. This study employs SSL models trained by predicting multiple fixed targets or masked frames. We propose a list of fixed targets to make the generated representation more efficient for detecting MDD and PTSD. Furthermore, we modify the hyper-parameters of the SSL encoder predicting fixed targets to generate global representations that capture varying temporal contexts. Both these innovations are noted to yield improved detection performances for considered mental disorders and exhibit task-agnostic traits. In the context of the SSL model predicting masked frames, the generated global representations are also noted to exhibit task-agnostic traits.
\end{abstract}


\section{Introduction}
\label{intro}
Recently, several studies have been reported focusing on the automatic detection of mental disorders using the recorded interactions of participants with a human interviewer/computer agent~\cite{DAIC14,ReviewMHD,Katharina_MDD_PTSD20}. In those setups, the data is collected preferably in audio/video mode. The main challenge in developing an effective detector of a mental disorder lies in finding appropriate feature representation of the audio/video data. In the early reported works, several acoustic and visual features and their combination have been explored for detecting a mental disorder. Those features are borrowed from different speech/visual-based pattern recognition tasks, predominantly from emotion recognition~\cite{4VQIS13,AVEC14,AVEC16,Katharina_MDD_PTSD20}. The quest for finding a tailored feature set for mental disorder detection is ongoing. Finding such a tailored feature set is challenging due to heterogeneity within a mental disorder~\cite{Hetro_MDs18}. In the last few years, several studies have explored different deep-learning architectures to generate a suitable latent representation for detecting mental disorders~\cite{DepAudioNet,MDD_AudText18,MDD_FVTC20}. Most of these studies mainly focus on the detection of a specific mental disorder. To the authors' knowledge, the exploration of a generic representation suitable for detecting multiple mental disorders has not yet been reported.
The justification behind the existence of a generic representation lies in the fact that multiple mental disorders are assigned common symptoms in the diagnostic and statistical manual of mental disorders~\cite{DSM5Book}. As a result, the audio/video data may encompass a variety of attributes associated with multiple disorders, and the same has been verified on the audio modality in our recent work~\cite{SPECOM23}.

The publicly available databases for developing a mental disorder detector are of limited sizes. In the literature, knowledge from other related domains with abundant resources has been leveraged to address the data scarcity in a domain, preferably through an unsupervised learning approach. The tasks in the former and latter domains are referred to as upstream and downstream tasks, respectively. The trained encoder in the upstream task is used as a feature extractor in the downstream task.
Recently, self-supervised learning (SSL), a branch of the unsupervised approach, has gained attention~\cite{SSL_review22}. The SSL employs an upstream task that conceptualizes the learning objective for training a model. The upstream tasks utilize pseudo labels generated from the input data itself. Based on the kind of pseudo labels utilized by the upstream task, the SSL is broadly categorized into three categories, namely \emph{constrastive}, \emph{predictive}, and \emph{generative}. In the contrastive approach, the encoder learns by comparing the positive and negative samples from the anchor sample. The objective is to maximize the similarity metrics between positive and anchor samples and minimize the similarity metrics between anchor and negative samples. The derivation of relevant negative samples is critical for training the encoder. Irrelevant negative samples may lead to the collapse of the encoder and thus make it necessary to use expert knowledge.
In the predictive approach, the targets for the training need to derive from a separate network. This approach is also known as the student-teacher approach. The teacher network provides the targets for training the student network. The use of teacher networks makes it computationally expensive. 
In the generative approach, the encoder learns by either reproducing the original input from the augmented one or generating a fixed target from the input. The augmentation techniques involve masking, shuffling, adding noise, etc. The fixed target(s) can be a predefined feature set(s). The generative approach also requires expert knowledge to select augmentation techniques or fixed targets.

Several studies have been reported that investigate the task-agnostic traits of SSL~\cite{MTSSL_Visual17,CPC_MM18,PASE19,TRILL20,GlobalRep22,Mockingjay20}. In those studies, the investigated downstream tasks mainly involved speech recognition, speaker recognition, emotion recognition, and/or a single mental disorder detection. 
However, a task-agnostic representation is yet to be explored for detecting multiple mental disorders to the best of the author's knowledge. In this study, we investigate the task-agnostic trait of SSL for detecting two correlated mental disorders, namely major depressive disorder (MDD) and post-traumatic stress disorder (PTSD), using audio and video modalities. For diversity purposes, the investigation is carried out by employing two recently proposed generative SSL models, namely PASE~\cite{PASE19} and AALBERT~\cite{AALBERT21}. The PASE and AALBERT models are based on architectures that involve predicting multiple fixed targets and masked frames, respectively. Both these models are originally trained to generate localized representations. However, a recent study reports that the global representations happen to be more effective than the local ones for MDD detection~\cite{GlobalRep22}. Motivated by that, in this study, the hyper-parameters (stride sizes of convolutional layers) of the PASE model are varied in such a way that it generates global representations that capture varying temporal contexts. For the AALBERT model, the global representation is obtained by applying average pooling in the temporal dimension of the generated localized representation.

The remaining paper is organized as follows. A brief description of considered SSL models in this study is provided in Section~\ref{SSLmodels}. Section~\ref{ExpDetails} illustrates the experimental details. The experimental results are provided in Section~\ref{Results}. Finally, the concluding remarks of the paper are provided in Section~\ref{conclusions}.

\section{SSL models}
\label{SSLmodels}
This section describes two kinds of generative SSL models utilized in this study. These SSL models differ in terms of upstream task being either multi-target prediction or masked prediction.
\subsection{Multi-target prediction}
\label{Sec:PASE}
In a recent study~\cite{PASE19}, the authors proposed a multi-target prediction-based SSL architecture and referred to it as PASE. This architecture is implemented to generate task-agnostic representation from raw speech. The downstream tasks involved speech recognition, speaker recognition, and emotion recognition. The PASE consists of a fully convolutional speech encoder, succeeded by seven parallel multilayer perceptron (MLP) workers that solve upstream tasks. These workers operate collaboratively to address various upstream tasks. The first layer of the encoder is made of SincNet~\cite{SincNet18}, which employs bandpass filters. The subsequent layers consist of conventional convolutional layers. An additional convolutional layer with a kernel size of 1, followed by a batch normalization layer, is used to project the latent representation to a predefined size of 100. The stride sizes of the convolutional layers are chosen in such a way that the input speech gets decimated in time by a factor of 160. Thus, it generates a representation of size 100 at every 10 msec, considering the sampling rate of 16 kHz. The encoded representation is further inputted to seven workers, which is defined as either regression or binary discrimination tasks. All regression workers, apart from the one that reproduces the input waveform, estimate a standard hand-crafted feature set. On the other hand, the binary discriminators are trained to discriminate positive and negative samples. The estimated losses from all workers are backpropagated to generate a representation suitable for all considered workers.

\begin{figure}[t]
\begin{center}
\centerline{\includegraphics[width=8cm,height=8cm,keepaspectratio]{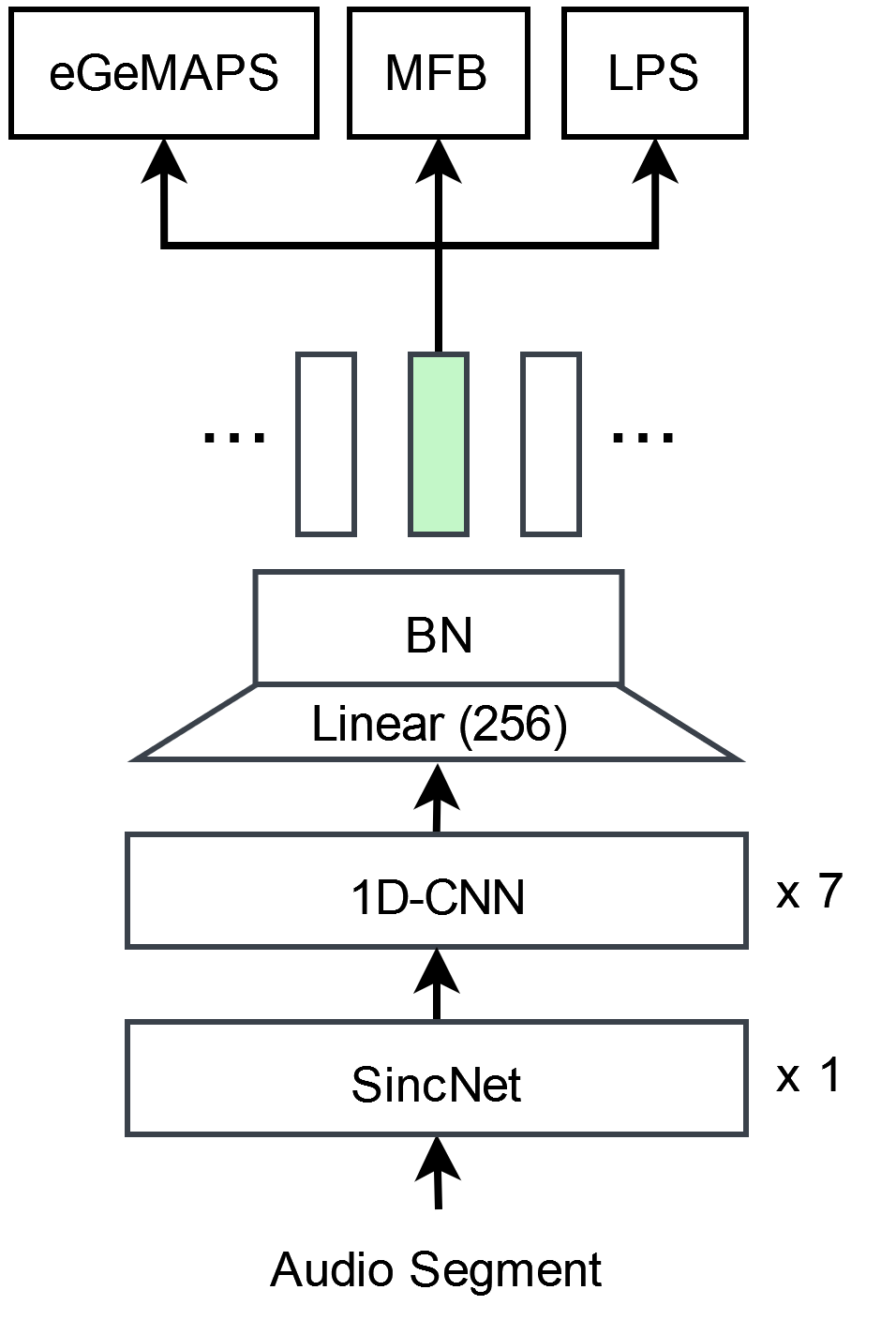}}
\caption{The schematic diagram of the multi-target prediction-based SSL model employed in this study.}
\label{PASE_mod}
\end{center}
\end{figure}

\begin{table}[b]
    \centering
    \caption{Set of stride sizes utilized in the convolutional layers in PASE and PASE-mod encoders.}
    \begin{tabular}{c c c}
    \hline
    \hline
      \textbf{Encoder} & \textbf{Decimation factor} & \textbf{Set of strides} \\
        \hline
       PASE &  160 & \{1, 10, 2, 1, 2, 1, 2, 2\}\\
       \hline
       PASE-mod &  160 & \{1, 10, 2, 1, 2, 1, 2, 2\}\\
       & 1.6k & \{1, 16, 1, 5, 2, 1, 2, 5\}\\
         & 8k & \{1, 16, 1, 5, 2, 5, 2, 5\}\\
         & 16k & \{1, 16, 2, 5, 2, 5, 2, 5\}\\
         & 32k & \{1, 16, 4, 5, 2, 5, 2, 5\}\\
         \hline
         \hline
    \end{tabular}
    \label{StridesConv}
\end{table}

For this study, we modified the default list of the workers proposed in~\cite{PASE19}. The modification is based on the hypothesis that a few workers in the default list may not be suitable for the detection of mental disorders of interest. The modified list of workers comprises extended Geneva minimalistic acoustic parameters (eGeMAPS), mel-filterbank (MFB) energies, and log power spectra (LPS). The eGeMAPS includes 23 low-level descriptors (LLDs) that are related to frequency, energy, and spectrum. The MFB comprises 40 mel-filterbank features. The LPS is computed using a Hamming window of 25 ms duration and a step size of 10 ms, with 1025 frequency bins per time step. Additionally, we increased the size of the representation from 100 to 256. Figure~\ref{PASE_mod} shows the schematic diagram of the PASE encoder with the modified list of the workers, and it is referred to as PASE-mod in the remaining text. 
As we discussed in Section~\ref{intro}, the inclusion of large context (i.e., global representation) is found to be advantageous in the detection of MDD~\cite{GlobalRep22}. Motivated by that, in this study, we generated global representation by adjusting the stride sizes of the seven convolutional layers following the SincNet layer in the PASE-mod encoder. The adjustment in the stride sizes is done in such a way that the resultant decimation factor increases to more than 160. Table~\ref{StridesConv} shows the emulated decimation factors and the corresponding set of stride sizes utilized in PASE and PASE-mod encoders. The kernel size and number of filters in each convolutional layer are kept identical to those mentioned in~\cite{PASE19}.

\subsection{Masked prediction}
\label{AALBERT}
\begin{figure}[ht]
        \centering
 \includegraphics[width=8cm,height=8cm,keepaspectratio]{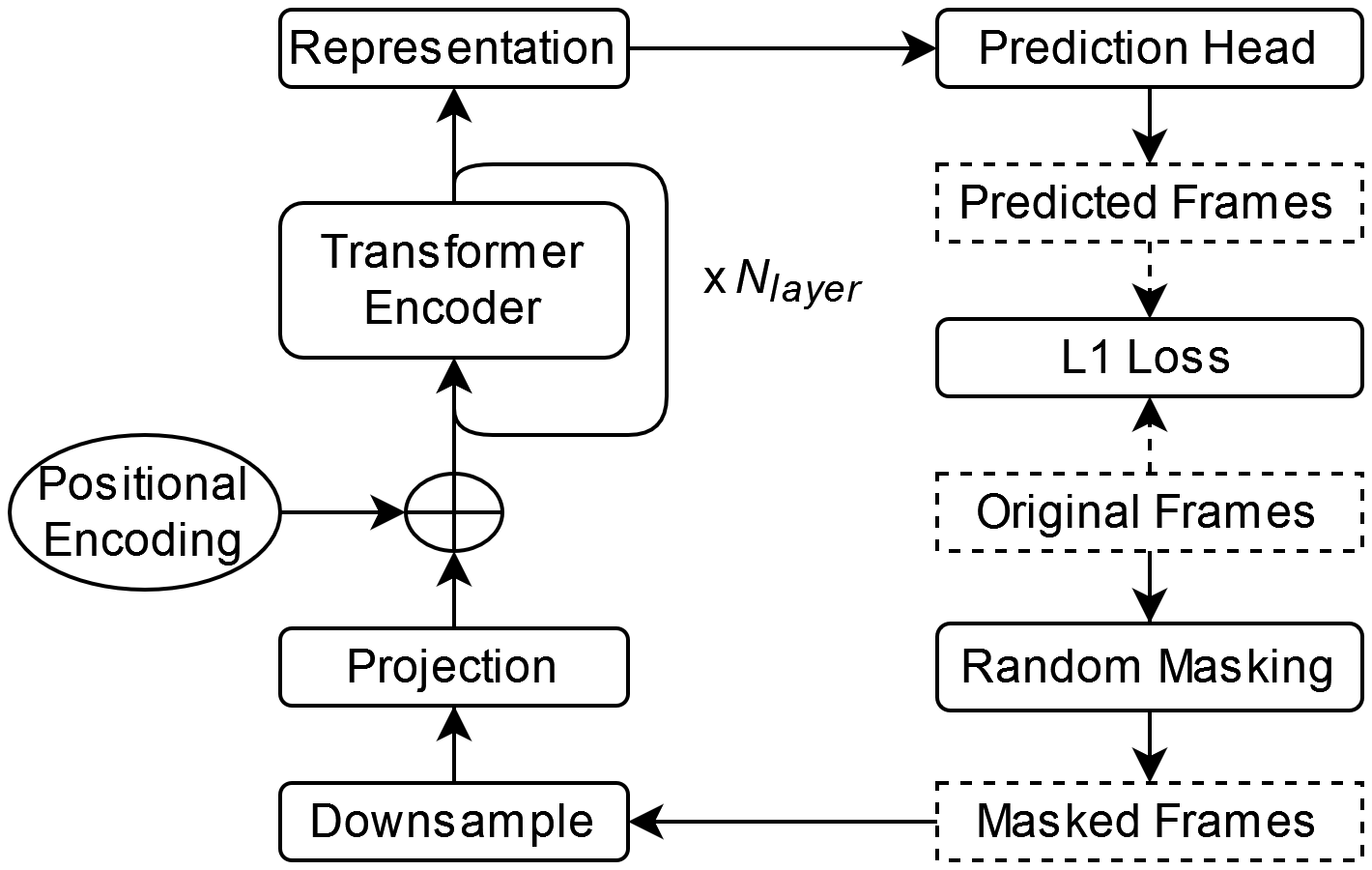}
        \caption{\footnotesize The schematic diagram of masked frames prediction-based SSL model employed in this study.}
        \label{fig:AALBERT}
    \end{figure}

In a recent study~\cite{AALBERT21}, the authors proposed a masked prediction-based SSL architecture and referred to it as AALBERT. The architecture comprises transformer layers with shared parameters. The upstream task of the model is to predict the masked frames. Figure~\ref{fig:AALBERT} shows the schematic diagram of the AALBERT model. The model was originally proposed for the audio modality, which takes the melspectrogram as the input. For this study, we utilized the model for video modality. The model is inputted with a segment comprising 147-dimensional feature vectors. A feature vector is the combination of action units (AUs), head pose (HP), eye gaze (EG), and their first- and second-order derivatives. Where AUs is a predefined set of human facial expressions. HP describes a participant's head position and orientation. EG captures the direction of the gaze of a participant's eyes. In this study, all model parameters and utilized strategies are kept identical to those mentioned in~\cite{AALBERT21}.

\section{Experimental Details}
\label{ExpDetails}
\subsection{Datasets}
\label{datasets}
For developing the MDD/PTSD detector, we have utilized the Distress Analysis Interview Corpus Wizard-of-Oz (DAIC-WOZ) dataset~\cite{DAIC14}. It comprises recorded sessions, in audio and video modes, of 189 participants interacting with a virtual interviewer~\cite{SimSensei14}. The recorded audio/video data of each participant are labeled based on the scores obtained from the Patient Health Questionnaire (PHQ-8)~\cite{PHQ8} and the PTSD Checklist -- Civilian version (PCL-C)~\cite{PCLC1998} questionnaires for MDD and PTSD detection, respectively. Instead of the raw video, only extracted features from the OpenFace toolbox~\cite{OPENFACE} are provided for ethical constraints. The dataset is further divided into training, development, and test partitions comprising data from 107, 35, and 47 participants, respectively. In order to use the recorded audio files for the experimentation, the embedded voice of the virtual interviewer is usually removed as it does not carry any relevant information for the intended tasks. For the same, the time stamps provided in the transcripts are utilized.

For training the considered SSL models, two distinct datasets, namely Interactive Emotional Dyadic Motion Capture (IEMOCAP)~\cite{IEMOCAP} and LibriSpeech~\cite{LibriSpeech}, along with the training partition of the DAIC-WOZ are utilized. The IEMOCAP comprises approximately 12 hours of data collected from 10 participants in audio and video mode during a dyadic conversation in five separate sessions. All the participants are actors who elicited distinct emotions during the scripted and improvised conversations. The dataset comprises labels of both discrete emotions and affective dimensions. The LibriSpeech, a popular dataset for speech recognition, comprises audio recordings from 2484 speakers. 

\subsection{Encoders and detectors}
\label{ExpEncDec}
\subsubsection{PASE/PASE-mod encoder}
The authors in~\cite{PASE19} trained the PASE encoder up to a predefined number of epochs, assuming that the validation losses have reached a plateau for all the workers. The trained encoder is then used as the feature extractor. In contrast, to introduce diversity, we used five different feature extractors obtained by training the encoder up to epochs ranging from 80 to 120 in step 10. Following the experimental setup given in~\cite{PASE19}, a subset of the LibriSpeech dataset is generated by randomly extracting a 15-sec segment for each of the 2484 speakers. The resultant subset, which comprises about 10 hours of data, is used for training the PASE and PASE-mod encoders. Furthermore, the PASE-mod encoder is also trained separately on the training partition of the DAIC-WOZ and the first three sessions of the IEMOCAP.

\subsubsection{AALBERT encoder}
\label{ExpEncAALBERT}
In this study, the AALBERT encoder is trained on video modality as discussed in Section~\ref{AALBERT}. For training the encoder, we experimented with different input segment lengths (4-sec and 10-sec), hop sizes (30\% and 100\%), and a number of transformer layers (3, 6, 9, and 12). The encoder is trained using the training partition of DAIC-WOZ, all five sessions of IEMOCAP, and their combination. All the model and training configurations are kept identical to those mentioned in~\cite{AALBERT21} except the model is trained for 100k steps. The feature extractor corresponds to the encoder with the lowest training loss. The generated localized representation for an input segment comprises 768-dimensional vectors. To generate a global representation, we applied average pooling along the temporal dimension of the localized representation. Thus, the global representation for each input segment is a 768-dimensional vector.

\subsubsection{Detectors}
For creating an MDD/PTSD detector, the generated representation by the encoder is subsequently inputted to an MLP classifier with a single hidden layer comprising 256 nodes. The training partition of DAIC-WOZ is significantly imbalanced between positive and negative classes. For addressing this class imbalance issue, the authors in~\cite{DepAudioNet} have used random sampling. Following that, in this study, we train the classifier five times on randomly generated training subsets, and the final outcome is obtained by averaging the probabilities predicted by those. The segment length for audio modality is set to 4 sec. Meanwhile, both 4-sec and 10-sec segments are used for video modality. The classifier is validated and tested on the development and test partitions of the DAIC-WOZ, respectively.

\subsection{Baselines}
\label{baseline}
\begin{figure}[ht]
    \centering
\includegraphics[width=5cm,height=5cm,keepaspectratio]{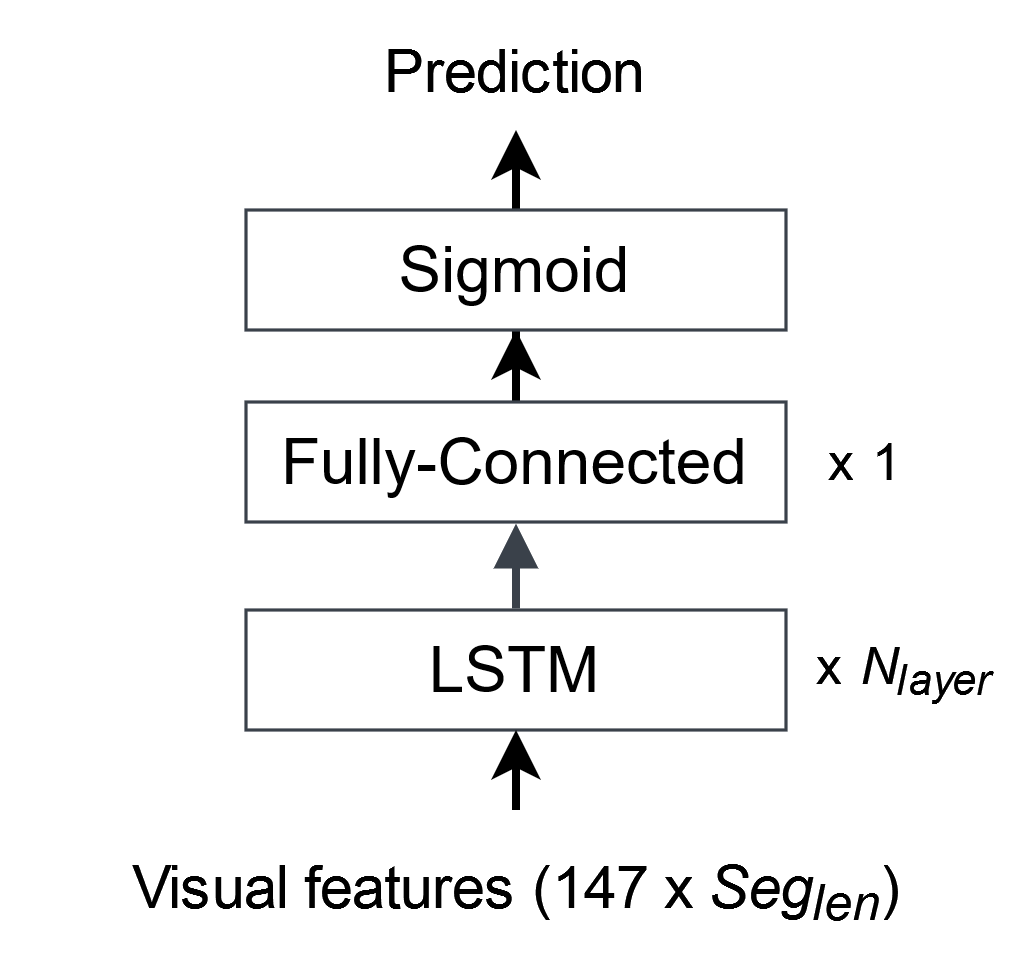}
\caption{The schematic diagram of the deep learning model architecture employed for video modality. $N_{layer}$ and $Seg_{len}$ denote the number of layers and segment length, respectively.}
\label{fig:LSTM_vis}
\end{figure}

To the best of our knowledge, the detection performance of PTSD on the DAIC-WOZ dataset is yet to be reported. For evaluating the efficacy of considered SSL models, we referred to the baselines created in our recent work~\cite{SPECOM23} for MDD and PTSD detection on audio modality. In that work, we employed DepAudioNet~\cite{DepAudioNet} and raw audio~\cite{GenderBiasDepression} architectures and experimented with varying the number of convolutional layers and the rate of decay of the learning rate. For MDD detection, the best performance in terms of macro-averaging F1-score was found to be 0.401 and 0.526 for DepaudioNet and raw audio, respectively. At the same time, those for PTSD detection were found to be 0.537 and 0.496.
For a mental disorder (MDD/PTSD), the absolute difference between the best detection performance and the detection performance on
the model that provided the best detection for the counterpart mental disorder (PTSD/MDD) is significantly high. The absolute difference is noted to be 0.125 and 0.041 for MDD and PTSD, respectively. These outcomes indicate that obtaining task-agnostic traits for detecting MDD and PTSD on supervised learning-based models is quite difficult.
For video modality, we employed multiple models incorporating varying numbers of long short-term memory (LSTM) layers shown in Figure~\ref{fig:LSTM_vis}. The inputs are identical to those used for training the AALBERT encoder and are described in Section~\ref{AALBERT}.

\section{Results}
\label{Results}
The primary objective of this study is to investigate the task-agnostic traits of SSL for detecting correlated mental disorders in audio and video modalities. For the same, we focus on MDD and PTSD. The investigation utilizes two generative SSL encoders trained via multi-task prediction (PASE) and masked prediction (AALBERT). Both encoders are trained on distinct datasets and are tuned to generate global representations. The generated global representations from audio and video data are utilized for the detection of MDD and PTSD, and their task-agnostic traits are analyzed. The reported MDD and PTSD detection performances in this study are in terms of the F1-score of positive class (PC), negative class (NC), and their average on the test partition of the DAIC-WOZ dataset. The considered deep learning models are realized using PyTorch modules running on NVIDIA RTX A6000 GPU.

\begin{table}[b]
\setlength\tabcolsep{4pt}
\begin{center}
\caption{MDD and PTSD detection performances in terms of F1-score (mean $\pm$ standard deviation) on the test partition of the DAIC-WOZ. The detection performances are evaluated for encoder being trained on distinct datasets. }
\begin{tabular}{c c c c}
\hline
\hline
\multicolumn{4}{c}{\textbf{MDD Detection}}\\
\hline
\hline
\textbf{Encoder (Dataset)} & \textbf{PC} & \textbf{NC} & \textbf{Average}\\
\hline
PASE (LibriSpeech) & $0.147 \pm  0.10$ & $0.772 \pm 0.03$ & $0.460 \pm 0.04$\\
PASE-mod (LibriSpeech) & $0.224 \pm 0.04$ & $0.796 \pm 0.01$ & $0.509 \pm 0.02$\\
PASE-mod (DAIC-WOZ) & $0.203 \pm 0.01$ & $0.787 \pm 0.02$ & $0.495 \pm 0.02$\\
PASE-mod (IEMOCAP) & $0.306 \pm 0.05$ & $0.767 \pm 0.03$ & $\textbf{0.536} \pm 0.03$\\

\hline
\hline
\multicolumn{4}{c}{\textbf{PTSD Detection}}\\
\hline
\hline
\textbf{Encoder (Dataset)} & \textbf{PC} & \textbf{NC} & \textbf{Average}\\
\hline
PASE (LibriSpeech) & $0.211 \pm  0.06$ & $0.724 \pm 0.01$ & $0.467 \pm 0.03$\\
PASE-mod (LibriSpeech) & $0.179 \pm 0.01$ & $0.743 \pm 0.01$ & $0.461 \pm 0.01$\\
PASE-mod (DAIC-WOZ) & $0.385 \pm 0.05$ & $0.792 \pm 0.01$ & $\textbf{0.588} \pm 0.03$\\
PASE-mod (IEMOCAP) & $0.348 \pm 0.04$ & $0.729 \pm 0.02$ & $0.538 \pm 0.03$\\
\hline
\hline
\end{tabular}
\label{tableLocalizedPASE}
\end{center}
\end{table}

\begin{table*}
\begin{center}
\caption{MDD and PTSD detection performances in terms of F1-score (mean $\pm$ standard deviation) on the test partition of the DAIC-WOZ. The detection performances are on the global representations generated at different decimation factors.}
\begin{tabular}{c c c c c c c}
\hline
\hline
\multicolumn{7}{c}{\textbf{MDD Detection}}\\
\hline
\hline
\textbf{Decimation factor} & \multicolumn{3}{c}{\textbf{DAIC-WOZ}} & \multicolumn{3}{c}{\textbf{IEMOCAP}}\\
\cmidrule(lr){2-4}
\cmidrule(lr){5-7}
 & \textbf{PC} & \textbf{NC} & \textbf{Average} & \textbf{PC} & \textbf{NC} & \textbf{Average}\\
\hline
Baseline & $0.203 \pm 0.01$ & $0.787 \pm 0.02$ & $0.495 \pm 0.02$ & $0.306 \pm 0.05$ & $0.767 \pm 0.03$ & $ 0.536 \pm 0.03$\\
\hline
1.6k & $0.243 \pm 0.04$ & $0.678 \pm 0.03$ & $0.461 \pm 0.03$ & $0.362 \pm  0.02$ & $0.703 \pm 0.03$ & $0.533 \pm 0.02$\\
8k & $0.328 \pm 0.01$ & $0.764 \pm 0.02$ & $0.546 \pm 0.01$ & $0.436 \pm  0.04$ & $0.695 \pm 0.02$ & $\textcolor{red}{0.565} \pm 0.03$\\
16k & $0.132 \pm 0.10$ & $0.814 \pm 0.01$ & $0.473 \pm 0.05$ & $0.344 \pm  0.06$ & $0.770 \pm 0.03$ & $0.557 \pm 0.04$\\
32k & $0.406 \pm 0.04$ & $0.765 \pm 0.02$ & $\textbf{0.586} \pm 0.03$ & $0.179 \pm  0.04$ & $0.778 \pm 0.02$ & $0.478 \pm 0.02$\\

\hline
\hline
\multicolumn{7}{c}{\textbf{PTSD Detection}}\\
\hline
\hline
\textbf{Decimation factor} & \multicolumn{3}{c}{\textbf{DAIC-WOZ}} & \multicolumn{3}{c}{\textbf{IEMOCAP}}\\
\cmidrule(lr){2-4}
\cmidrule(lr){5-7}
 & \textbf{PC} & \textbf{NC} & \textbf{Average} & \textbf{PC} & \textbf{NC} & \textbf{Average}\\
\hline
Baseline & $0.385 \pm 0.05$ & $0.792 \pm 0.01$ & $0.588 \pm 0.03$ & $0.348 \pm 0.04$ & $0.729 \pm 0.02$ & $0.538 \pm 0.03$\\
\hline
1.6k & $0.355 \pm 0.122$ & $0.771 \pm 0.02$ & $0.563 \pm 0.06$ & $0.321 \pm  0.07$ & $0.710 \pm 0.03$ & $0.515 \pm 0.03$\\
8k & $0.249 \pm 0.01$ & $0.739 \pm 0.02$ & $0.494 \pm 0.02$ & $0.466 \pm  0.02$ & $0.789 \pm 0.02$ & $\textbf{0.628} \pm 0.02$\\
16k & $0.188 \pm 0.01$ & $0.760 \pm 0.03$ & $0.474 \pm 0.02$ & $0.293 \pm  0.08$ & $0.716 \pm 0.02$ & $0.504 \pm 0.04$\\
32k & $0.301 \pm 0.03$ & $0.744 \pm 0.01$ & $0.523 \pm 0.02$ & $0.277 \pm  0.04$ & $0.723 \pm 0.01$ & $0.500 \pm 0.02$\\

\hline
\hline
\end{tabular}
\label{tableGloablePASE}
\end{center}
\end{table*}

\subsection{Audio modality}
\label{res_audio}
Table~\ref{tableLocalizedPASE} shows the detection performances of MDD and PTSD. The PASE and PASE-mod encoders are trained to generate representations with a decimation factor of 160, referred to as localized representations in the remainder of the text. The obtained representation is flattened before being inputted into the detector. Upon utilizing these representations for developing MDD/PTSD detectors, poor training accuracies are obtained. To address the same, we applied average pooling on ten consecutive representations and then flattened.
In Table~\ref{tableLocalizedPASE}, for each case, the listed mean and standard deviation of the detection performances yielded by the detectors that are trained separately on the representations obtained from five different feature extractors as discussed in Section~\ref{ExpEncDec}. The table shows that upon only modifying the list of workers while keeping the rest of the experimental setting identical, MDD detection performance has relatively improved by 10.65\%. However, PTSD detection performances on both encoders are quite similar.
Furthermore, the PASE-mod encoder is trained separately on the training partition of DAIC-WOZ and IEMOCAP. It can be observed from the table that the PASE-mod encoder trained on IEMOCAP and on the training partition of the DAIC-WOZ provided the best detection performance for MDD and PTSD, respectively. The relative increment compared to the PASE encoder trained on LibriSpeech is found to be 16.52\% and 25.91\%, respectively, for MDD and PTSD. However, the yielded best detection performances for MDD and PTSD are on different sets of representations. The detection performance of the counterpart mental disorder on the same set of representations has relatively degraded. The relative decrement is found to be 7.65\% and 8.50\%, respectively, for MDD and PTSD.

For the further investigation of task-agnostic representation, we generated global representations with varying decimation by assigning a different set of stride sizes to the PASE-mod encoder as described in Section~\ref{Sec:PASE}. The encoder is trained separately using the training partition of DAIC-WOZ and IEMOCAP. Table~\ref{tableGloablePASE} shows the detection performances of MDD and PTSD detectors that are trained on the different global representations. The best detection performances for MDD and PTSD found in Table~\ref{tableLocalizedPASE} are also included as baselines for contrast purposes. It can be observed from Table~\ref{tableGloablePASE} that the encoder trained on the training partition of DAIC-WOZ with a decimation factor of 32k provides the best MDD detection. On the other hand, the encoder trained on IEMOCAP with a decimation factor of 16k provides the best PTSD detection. As the yielded best detection performances for MDD and PTSD are on different sets of representations, the representations generated by the encoder trained on IEMOCAP with an 8k decimation factor would be most suitable for task-agnostic purposes. Given that it produces the best results for detecting PTSD, its performance in detecting MDD closely aligns with the best results obtained. Also, the yielded detection performance outperformed the corresponding baseline. The relative increment in the detection performance for MDD and PTSD detection is found to be 5.41\% and 16.73\%, respectively.

\begin{table}[b]
\centering
\caption{Comparing  MDD and PTSD detection performances on task-agnostic representation with corresponding baselines in terms of macro-averaging F1-score (macro-F1).}
\begin{tabular}{c c c}
\hline
\hline
\textbf{Mental Disorder} & \textbf{Model} & \textbf{macro-F1}\\
\hline
MDD & DepAudioNet & 0.401\\
& Raw Audio & 0.526\\
\cmidrule(lr){2-3}
 & PASE-mod & \textbf{0.565}\\
\hline
 PTSD & DepAudioNet & 0.537\\
& Raw Audio & 0.496\\
\cmidrule(lr){2-3}
 & PASE-mod & \textbf{0.628}\\

\hline
\hline
\end{tabular}
\label{tableAudio}
\end{table}

For validating the detection performances of MDD and PTSD on task-agnostic representations, we compared them with the baselines described in Section~\ref{baseline}. It can be noted from Table~\ref{tableAudio} that the detection performances on task-agnostic representations generated by PASE-mod encoder outperformed the corresponding baselines.

\begin{table}[t]
\setlength\tabcolsep{4.5pt}
\centering
\caption{MDD and PTSD detection performances in terms of F1-score on the test partition of the DAIC-WOZ. The considered deep learning models comprise different numbers of LSTM layers and are inputted with two different segment lengths.}
\begin{tabular}{c c c c c c c c}
\hline
\hline
\textbf{Disorder} & \textbf{\#Layers} & \multicolumn{3}{c}{\textbf{4-sec}} & \multicolumn{3}{c}{\textbf{10-sec}}\\
\cmidrule(lr){3-5}
\cmidrule(lr){6-8}
 &  & \textbf{PC} & \textbf{NC} & \textbf{Avg.} & \textbf{PC} & \textbf{NC} & \textbf{Avg.}\\
\hline
MDD & 1 & 0.345 & 0.708 & 0.526 & 0.585 & 0.679 & \textbf{0.632}\\
& 2 & 0.400 & 0.719 & 0.559 & 0.409 & 0.480 & 0.445\\
& 3 & 0.357 & 0.727 & 0.542 & 0.578 & 0.612 & 0.595\\
& 4 & 0.296 & 0.716 & 0.506 & 0.558 & 0.628 & 0.593\\
& 5 & 0.211 & 0.800 & 0.505 & 0.578 & 0.612 & 0.595\\
\hline
PTSD & 1 & 0.483 & 0.769 & \textbf{0.626} & 0.462 & 0.618 & 0.540\\
& 2 & 0.400 & 0.783 & 0.591 & 0.571 & 0.654 & 0.613\\
& 3 & 0.273 & 0.778 & 0.525 & 0.410 & 0.582 & 0.496\\
& 4 & 0.200 & 0.784 & 0.492 & 0.462 & 0.618 & 0.540\\
& 5 & 0.211 & 0.800 & 0.505 & 0.267 & 0.656 & 0.462\\
\hline
\hline
\end{tabular}
\label{tableBaselineAALBERT}
\end{table}

\subsection{Video modality}
\label{res_video}
 In this study, we utilized the AALBERT encoder for video modality. The input to the encoder is either a 4-sec or 10-sec segment comprising 147-dimensional feature vectors (ref. Section~\ref{AALBERT}). For creating the baseline, we implemented the LSTM model with different numbers of layers as discussed in Section~\ref{baseline}. The LSTM models are trained on the 2-dimensional feature representation identical to those used for training the AALBERT encoder. Table~\ref{tableBaselineAALBERT} shows the detection performances of MDD and PTSD for the LSTM models. It can be observed from the table that the best detection performance for MDD and PTSD is yielded by the one-layered LSTM model trained on segments of length 10 sec and 4 sec, respectively. 
For a mental disorder (MDD/PTSD), the absolute difference between the best detection performance and the detection performance on the model that provided the best detection for the counterpart mental disorder (PTSD/MDD) is significantly high. The absolute difference is noted to be 0.106 and 0.086 for MDD and PTSD, respectively. These outcomes suggest that obtaining task-agnostic traits for detecting MDD and PTSD on supervised learning-based models is quite difficult. This observation is consistent with the one found for audio modality in Section~\ref{baseline}.
Table~\ref{tableAALBERT} shows the detection performances of MDD and PTSD detectors that are trained on the global representation generated by the AALBERT encoder. The encoder is trained separately on the training partition of DAIC-WOZ and IEMOCAP. The baseline in the table corresponds to the best detection performances found in Table~\ref{tableBaselineAALBERT}. For brevity, we only listed the best detection performance yielded for MDD/PTSD across considered hyper-parameters, described in Section~\ref{ExpEncAALBERT}. For evaluating the task-agnostic traits, we also listed the detection performance of a mental disorder on the representation that provided the best detection performance for another mental disorder. It is found that the best detection performances are yielded by the encoder with six transformer layers trained on 4-sec segments and 30\% hop size. It can be noted from the table that the AALBERT encoder trained on the training partition of DAIC-WOZ and IEMOCAP outperformed the corresponding baseline for MDD and PTSD detection, respectively. The relative increment is found to be 8.70\% and 8.79\%, respectively, for MDD and PTSD detection. However, the detection performance of another mental disorder on the same representation is significantly degraded. The relative decrement compared to the baseline is found to be 5.27\% and 9.18\%, respectively, for PTSD and MDD.

\begin{table}[t]
\centering
   \caption{MDD and PTSD detection performances in terms of F1-score on the test partition of the DAIC-WOZ. The AALBERT encoder is trained separately on the training partition of DAIC-WOZ and IEMOCAP.}
\begin{tabular}{c c c c c}
\hline
\hline
\textbf{Dataset} & \textbf{Mental Disorder} & \textbf{PC} & \textbf{NC} & \textbf{Avg.}\\
\hline
Baseline & MDD & 0.585 & 0.679 & 0.632\\
 & PTSD & 0.483 & 0.769 & 0.626\\
\hline
DAIC-WOZ & MDD & 0.581 & 0.794 & \textbf{0.687}\\
 & PTSD & 0.429 & 0.758 & 0.593\\
 \hline
IEMOCAP & MDD & 0.438 & 0.710 & 0.574\\
& PTSD & 0.539 & 0.824 & \textbf{0.681}\\

\hline
\hline
\end{tabular}
\label{tableAALBERT}
\end{table}

\begin{table}[b]
\centering
\caption{MDD and PTSD detection performances in terms of F1-score on the test partition of the DAIC-WOZ for AALBERT encoder having different numbers of transformer layers. The encoder is trained on the combination of the training partition of DAIC-WOZ and IEMOCAP.}
\begin{tabular}{c c c c c}
\hline
\hline
\textbf{\#Layers} & \textbf{Mental Disorder} & \textbf{PC} & \textbf{NC} & \textbf{Avg.}\\
\hline
Baseline & MDD & 0.585 & 0.679 & 0.632\\
 & PTSD & 0.483 & 0.769 & 0.626\\
\hline
3 & MDD & 0.487 & 0.667 & 0.577\\
 & PTSD & 0.400 & 0.719 & 0.559\\
 \hline
6 & MDD & 0.533 & 0.781 & 0.657\\
& PTSD & 0.296 & 0.716 & 0.506\\
\hline
9 & MDD & 0.529 & 0.733 & 0.631\\
& PTSD & 0.500 & 0.788 & \textbf{0.644}\\
\hline
12 & MDD & 0.588 & 0.767 & \textbf{0.678}\\
& PTSD & 0.445 & 0.776 & 0.610\\
\hline
\hline
\end{tabular}
\label{tableCombinedDatasetsAALBERT}
\end{table}

For further investigation of task-agnostic representation, we trained the AALBERT encoder on the combined dataset from the training partition of DAIC-WOZ and IEMOCAP. Table~\ref{tableCombinedDatasetsAALBERT} shows the MDD and PTSD detection performances of the detectors trained on the global representation generated by the AALBERT encoder with different numbers of transformer layers. We solely reported the detection performances on representations generated by the encoder trained on segments of 4-sec length with a 30\% hop size, as it outperformed other hyper-parameter combinations. It can be noted from the table that the encoder with 9 and 12 transformer layers provides the best detection for PTSD and MDD, respectively.
With 9 transformer layers, the PTSD detection performance has increased by 2.88\% relative to the baseline, while the MDD detection has decreased relatively by 0.16\%. On the other hand, with 12 transformer layers, the MDD detection performance has relatively increased by 7.28\%, but the PTSD detection has decreased relatively by 2.56\% compared to their baselines. The relative decrements are not very significant.
Thus, the global representation generated by the AALBERT encoder, trained on the combined dataset from the training partition of DAIC-WOZ and IEMOCAP, with either 9 or 12 transformer layers, can be used for task-agnostic purposes.

\section{Conclusions}
\label{conclusions}
This study investigates the task-agnostic traits of representations derived through SSL for detecting correlated mental disorders using both audio and video modalities. The investigation is carried out using two recently proposed generative SSL models (PASE and AALBERT) for MDD and PTSD detection tasks. The results show that the detection performance yielded on the global representations generated by PASE-mod and AALBERT outperformed the created baselines using supervised deep learning models.
Thus, the global representations generated by the said SSL models can be used for task-agnostic purposes.

This study is limited to two correlated mental disorders. It would be interesting to explore various other correlated mental disorders. Replicating this study using different SSL models would be another direction for future work.

\bibliographystyle{IEEEtran}

\bibliography{Bibfile}

\end{document}